\documentclass[sigconf]{acmart}
\usepackage{multirow}
\usepackage{color}
\AtBeginDocument{%
  \providecommand\BibTeX{{%
    \normalfont B\kern-0.5em{\scshape i\kern-0.25em b}\kern-0.8em\TeX}}}

\begin{document}

\title{Text-Guided Neural Image Inpainting}

\author{ Lisai Zhang$^{1}$, \ Qingcai Chen*$^{1,2}$, \ Baotian Hu$^{1}$, \ Shuoran Jiang$^{1}$}
\affiliation{\institution{$^{1}$Shenzhen Chinese Calligraphy Digital Simulation Engineering Laboratory, Harbin Institute of Technology, Shenzhen}}
\affiliation{\institution{$^{2}$Peng Cheng Laboratory, Shenzhen, China }}
\affiliation{\institution{LisaiZhang@foxmail.com, \ \ qingcai.chen@hit.edu.cn*, \ \ \{baotianchina,shaunbysn\}@gmail.com}}

\renewcommand{\shortauthors}{Zhang, et al.}
\renewcommand{\authors}{Lisai Zhang, Qingcai Chen, Baotian Hu, Shuoran Jiang}
\begin{abstract}
Image inpainting task requires filling the corrupted image with contents coherent with the context. This research field has achieved promising progress by using neural image inpainting methods. Nevertheless, there is still a critical challenge in guessing the missed content with only the context pixels. The goal of this paper is to fill the semantic information in corrupted images according to the provided descriptive text. Unique from existing text-guided image generation works, the inpainting models are required to compare the semantic content of the given text and the remaining part of the image, then find out the semantic content that should be filled for missing part. To fulfill such a task, we propose a novel inpainting model named Text-Guided Dual Attention Inpainting Network (TDANet). Firstly, a dual multimodal attention mechanism is designed to extract the explicit semantic information about the corrupted regions, which is done by comparing the descriptive text and complementary image areas through reciprocal attention. Secondly, an image-text matching loss is applied to maximize the semantic similarity of the generated image and the text. Experiments are conducted on two open datasets. Results show that the proposed TDANet model reaches new state-of-the-art on both quantitative and qualitative measures. Result analysis suggests that the generated images are consistent with the guidance text, enabling the generation of various results by providing different descriptions. Codes are available at \textcolor[rgb]{1.00,0.00,1.00}{\texttt{\url{https://github.com/idealwhite/TDANet}}}.
\renewcommand{\thefootnote}{\fnsymbol{footnote}} 
\footnotetext{$^\ast$Corresponding author.}
\end{abstract}


\begin{teaserfigure}
	\vspace{-5pt}
	\begin{center}
		\setlength{\tabcolsep}{0.15em}
		\begin{tabular}{cccccl}
			\centering
			\includegraphics[width=2.6cm, height=2.6cm]{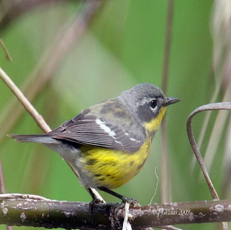} & 
			\includegraphics[width=2.6cm, height=2.6cm]{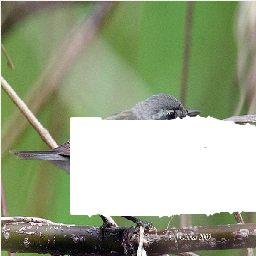} &\includegraphics[width=2.6cm, height=2.6cm]{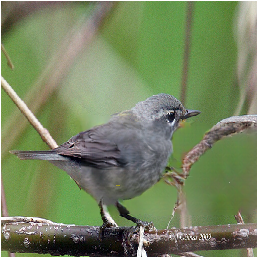} &
			\includegraphics[width=2.6cm, height=2.6cm]{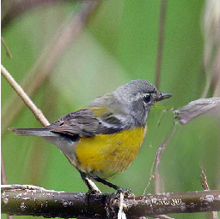} & 
			\includegraphics[width=2.6cm, height=2.6cm]{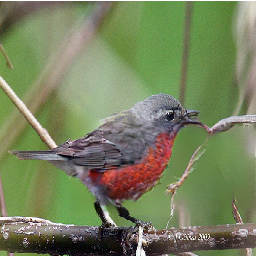}& 
			\includegraphics[width=2.6cm, height=2.6cm]{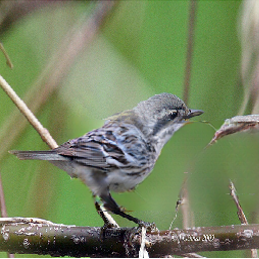} \\
			\small Origin & \small Corrupted & \small Without Text & \small \textbf{Text-guided (a)} & \small \textbf{Text-guided (b)} & \multicolumn{1}{c}{\small \textbf{Text-guided (c)}} \\
			\multicolumn{6}{l}{ \textbf{Text (a)}: \textit{``This bird has gray head, black and white wings and yellow belly.''} \quad \textbf{Text (b)}: \textit{ ``This bird is gray in color, with red belly.''}} \\
			\multicolumn{6}{l}{ \textbf{Text (c)}: \textit{``A small bird with spotted blue wings, gray tail and head, and has white throat, breast, belly and white undertail.''}}
		\end{tabular}
	\end{center}
	\vspace{-10pt}
	\caption{Illustration of inpainting a unique area. (a) is guided inpainting case with the original image description, (b) and (c) produces different new contents while guided with altered texts. }
	\label{fig:demo}
\end{teaserfigure}

\maketitle

\section{Introduction}
Image inpainting is the task to generate visually realistic content in missing regions while keeping coherence~\cite{ma2019coarse}. It plays an important role in many digital image processing tasks, such as restoration of damaged paintings, photo editing, and image rendering~\cite{bertalmio2000image}. There have been many methods proposed for generating semantically coherent content, such as integrating context features~\cite{plu_14,2019_coherent}, enhance convolution layers~\cite{plu_24,2019_gate}. and pluralistic completion~\cite{2019_pluralistic} through probability framework. 

A common assumption for inpainting models is that the missing area should have  patterns similar to the remaining region. For example, diffusion-based~\cite{plu_5} and patch-based~\cite{coherent_21} models use the remaining image to recover the missed regions. These models produce high-quality images, but usually usually fail in complicated scenes such as unique masks on objects or large holes~\cite{plu_40}. In recent years, deep learning-based image inpainting methods have been presented to overcome this limitation~\cite{plu_40}. 
An encoder-decoder inpainting structure with Generative Adversarial Networks (GAN) was first proposed by Pathak et al.~\cite{plu_29}. Satoshi et al.~\cite{plu_14} further improved the ability to encode the corrupted image and achieved photo-realistic results by adopting dilated convolutions and proposing global and local discriminators. However, the model cannot achieve high image appearance quality when filling irregular holes. Later works on neural image inpainting have achieved credible generation quality on irregular masks from the perspective of improving convolution~\cite{2019_gate,ma2019coarse} and integrating context features \cite{2019_coherent}. Although the appearance quality is constantly improving, it is still a challenge to generate accurate content for unique areas with patterns different than the remaining region.

Unique areas are hard to fill because there could be many semantically different solutions consistent with surrounding pixels. One of the directions for solving this problem is generating a diverse set of inpainting results. For example, a recent model PICNet~\cite{2019_pluralistic} proposed a dual pipeline training architecture to learn distribution for the masked area. The model generates plausible pluralistic predictions for single masked input. However, the solution space for such methods is very large. Even if they generate all possible solutions, the desired result needs to be chosen from numerous outputs, which is not only time-consuming but also resource-intensive. Giving external guidance to inpainting models is another common approach to control the output and reduce the solution space. Some user-guided image inpainting approaches allow external guidance, such as a line \cite{2019_gate} and edges~\cite{2019_edgeconnect}, and exemplar~\cite{gate_20}. However, these models provide only simple graphics tips and lack semantic diversity. What is more, the quality of user-guided inpainting largely depends on the quality of guidance, but suitable exemplar or drawing is sometimes difficult to obtain. 

Since image content can be described as descriptive texts most of the time, it would be feasible for inpainting models to borrow information about the hole from text descriptions. For example, in Fig.~\ref{fig:demo}, the color of the bird's belly is hard to be inferred only according to the remaining parts, but given a text description, the inpainting model will have a clear target. Some research on using text guidance already exists for other tasks, such as image generation \cite{attn_20,attn_27,attn_36,attngan}, These text to image generation works mainly focus on generating a complete image but do not take into account the constraints of image context. Image manipulation researchers also explored changing image content according to the text guidance \cite{manipulation_nips,bowen_2}. These models can change the pixels within the whole image area. Compared to text to image synthesis and text-guided manipulation, the text-guided inpainting task has more strict requirements. It requires identifying semantics that is complementary to existing image content, and generating coherent pixels at a fixed position.

In the text-guided inpainting task, the relationship between description and image samples is one-to-many: a semantic describes various samples that have the same meaning. It is natural to model such a relationship as probability distribution through the CVAE-like framework~\cite{cvaetutorial}. However, CVAE has been proven to have a disadvantage of grossly underestimating variances in the image completion scenario~\cite{2019_pluralistic}, which can be solved through the dual probabilistic structure. To use this structure in the text-guided inpainting task, we need to extend it to support the multimodal condition.

In this paper, we move one step further along user-guided inpainting and propose a novel model that fills the holes with the guidance of descriptive text. 
We design a novel dual multimodal attention mechanism to exploit the text features about the masked region by comparing text with the corrupted image and its counterpart. Then an image-text matching loss is adapted to regularize the similarity between text and model output. Experiments are conducted on two image caption datasets. The object boxes are used as the mask to unique areas. The inpainting results are evaluated in both qualitative and quantitative ways. 

The main contributions of this paper are listed as follows:
\begin{itemize}
	\setlength\itemsep{0em}
	\item A text-guided dual attention model is proposed to complete image with the guidance of descriptive text. The combination of text and image provides richer semantics than only using the corrupted image.
	
	\item A novel inpainting scheme is presented that enables entering different texts to get pluralistic outputs. 
	
	\item New state-of-the-art performances are reached by the proposed model on two public benchmarks.
\end{itemize}


\begin{figure*}
	\centering
	\includegraphics[width=0.9\linewidth]{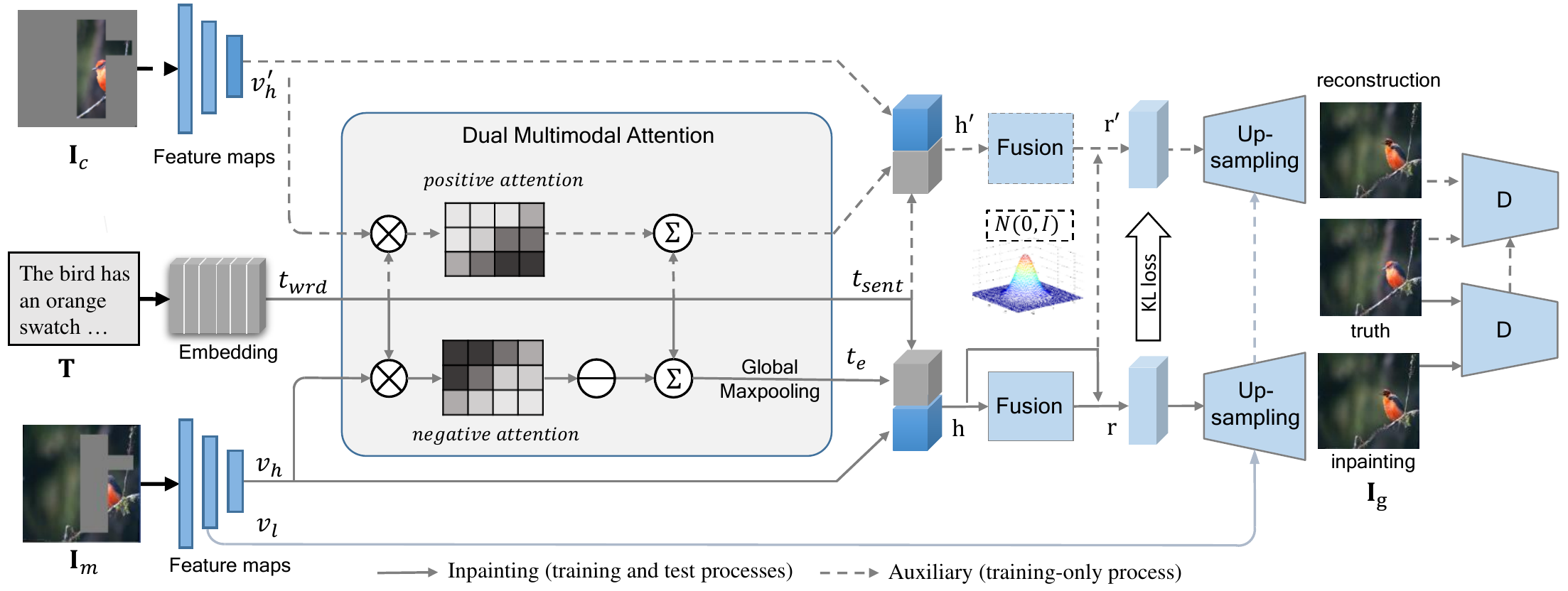}
	\caption{Architecture of the TDANet. The auxiliary path in dotted line is calculated only during training. The bold line is the inpainting path that infers the missing region from $\mathbf{I}_m$ and $\mathbf{T}$. The two paths share weights except in fusion modules.}
	\label{fig:model}
\end{figure*}

\section{Related Work}
\textbf{Image inpainting}
Traditional diffusion-based or patch-based approaches \cite{plu_5,coherent_21} fill missing regions by propagating neighbor information\cite{coherent_1,plu_38} or copying from similar patches~\cite{coherent_35,gate_11} of the background based on low-level features. These methods work well for surface textures synthesis but often fail on non-stationary textural images~\cite{2019_pluralistic}. 
To address the problem, Simakov et al. propose a bidirectional similarity synthesis approach \cite{plu_35} to better capture non-stationary information, but lead to high algorithm complexity. 
Recently, deep learning models are introduced to image inpainting that directly generates pixel values of the hole. Context encoders \cite{plu_29} use the encoder-decoder structure and conditional generative adversarial network to image inpainting tasks. Next, contextual attention  \cite{2018_plu_42_contextualattention} is proposed for capturing long-range spatial dependencies. To improve the performance on irregular masks, \cite{plu_24} proposes partial convolution  where the mask is updated, and convolution weights are re-normalized with layers. However, it has limitations to capture detailed information of the masked area in deep layers. For this, \cite{2019_gate} proposes to use gated convolution where a gate network learns the shape of mask in convolution. These approaches can produce only one result for each incomplete image. Thus, Zheng et al. \cite{2019_pluralistic} introduces a CVAE~\cite{plu_2} based pluralistic image completion approach. During inference, the model obtains the distribution for the input, then samples various representation vectors from the distribution, and feeds these vectors to the decoder to get a variety of outputs. Our model builds upon this dual training structure and explores how it can be used for user-guided image inpainting.

\textbf{User-guided Image inpainting}
Many user-guidance mechanisms are explored to enhance image inpainting systems, including dots or lines~\cite{gate_1}, structures~\cite{plu_13}, transformation or distortion~\cite{gate_30}, and exemplars~\cite{gate_10,gate_20}. Some methods are also based on patch match algorithms, but use external image sources~\cite{plu_7,gate_51} such as search engine~\cite{gate_43}. Recent advances in conditional generative networks benefit user-guided inpainting and synthesis. Wang et al. \cite{plu_39} proposes to synthesize high-resolution photo-realistic images from semantic label maps using conditional generative adversarial networks. \cite{2019_gate} extends this model to support user-guided inpainting with sketches. Compared with these schemes, image inpainting guided with text is challenging in two aspects: Firstly, image and text are heterogeneous; it is hard to transform image and text features to a shared space. What's more, the descriptive text usually contains redundant information, and the model must distinguish between the information about the corrupted region and the remaining parts.

\textbf{Text-guided Image Synthesis and Manipulation}
The development of text to image synthesis brings the possibility of generating images based on text prior. The task is different from text-guided image inpainting and directly generates an image from the text description.  Here we selectively review several related works. \cite{attn_20} first shows that the conditional GAN is capable of synthesizing plausible images from text descriptions. \cite{attn_36} stacks several GANs for text to image synthesis. However, their methods are conditioned on the global sentence vector, missing fine-grained word-level information. AttnGAN \cite{attngan} develops word and sentence level matching networks to generate fine-grained images from text. More recent works \cite{bowen_1,objectgan} exploit object and word-level information during synthesis through an attention mechanism. Compared with image synthesis task, the requirement for inpainting is more stringent: the generated content must coherent with remaining parts.
Some image manipulation work \cite{manipulation_nips,bowen_2} explored text guidance; the method automatically edits an image and changes its content according to the text description. In manipulation task, the model decides which pixels to change, but inpainting requires to generate content for fixed position.
Our research adopts the matching loss in AttnGAN to a new task, where the prior of generation is the combination of text and image, and generation target is a sub-region.

\section{Approach}
The proposed TDANet can be formulated as follows: given the masked input image $\mathbf{I}_m$ and descriptive text $\mathbf{T}$, the model outputs the target image $\mathbf{I}_g$.  The model uses the dual probabilistic structure and extend it to the multimodal condition. The overall structure of our model is shown in Figure~\ref{fig:model}. It composes of three components: Encoders for Image and Text, Dual multimodal Attention, and Inpainting Generation. 

In the auxiliary path, the input $\mathbf{I}_c$ is the masked region of the original image. We use $x'$ to denote the variable $x$ is now calculated in the auxiliary path.
\subsection{Preliminaries}
\label{sec.pre}
The DAMSM loss~\cite{attngan} is composed of pre-trained networks that match the similarity of image and texts. By maximizing similarity score, these pre-trained networks calculate gradients as discriminators that enforce the generated image to follow the text description. In detail, the word-image similarity is defined as Eq.~\ref{eq:d_word}
\begin{equation}
\label{eq:d_word}
S(I, T) = \log(\sum_{i=1}^{L-1}\exp(\gamma cos(I_{i},T_{i})))^\frac{1}{\gamma}
\end{equation}
where $I$ are the features of a generated image, $T$ is the word embeddings, and $L$ is the length of the sentence. For every batch of sentence-image pairs, the similarity score is computed as Eq.~\ref{eq:d_batch}.
\begin{equation}
\label{eq:d_batch}
P(I,T) = \frac{\exp(\gamma_2S(I_{i},T_{i}))}{\sum_{j=1}^{B}\exp(\gamma_2S(I_{i},T_{j}))}
\end{equation}
The word-image similarity could be optimized by minimizing the log probability of the score as Eq.~\ref{eq:d_score}.
\begin{equation}
\label{eq:d_score}
DAMSM_w = -\sum_{i=1}^{B}(\log P(I,T) + \log P(T,I))
\end{equation}
The sentence-image similarity $DAMSM_s$ is calculated in the same way, except for redefining Eq.~\ref{eq:d_word} as $S(I, T) =\cos(\bar{I},\bar{T})$. 

The short+long term attention~\cite{2019_pluralistic} is proposed to effectively reconstruct the unmasked areas in the inpainting model. The network computes a weight map between encoder (long term) and decoder (short term) features through a self-attention network, and then weights and sums these feature maps. The generator concatenates the weighted feature maps and reconstructs the unmasked areas.

\subsection{Encoders for Image and Text}
\label{sec.encoder}
In the image inpainting task, only the lost areas need to be inferred, and the remaining pixels could be reconstructed. We feed the input image to a 7-layer ResNet~\cite{resnet}, and extract features from different layers. The feature map of the top layer is taken as the high-level representation $v_h$; the output of the second last layer is used as low-level features $v_l$ . We use the high-level features to build the prior condition of inference and feed the low-level feature $v_l$ to the generator to reconstruct the unmasked areas. The image encoder in the auxiliary path shares the same weight with the one in the inpainting path.

As for the input sentence, a GRU network pretrained as~\cite{attngan} is used to compute a sequence of word representations $t_{wrd}$ and a sentence representation $t_{sent}$.  The hidden size of GRU network is 256. A sentence usually describes many parts of the image, so the key information about the missing region only exists in a subset of words. In order to use the word embedding to guide inpainting, the model must learn to extract this part of the information.

\subsection{Dual Multimodal Attention Mechanism}
\label{section:contrary}
\begin{figure}
	\centering
	\includegraphics[width=\linewidth]{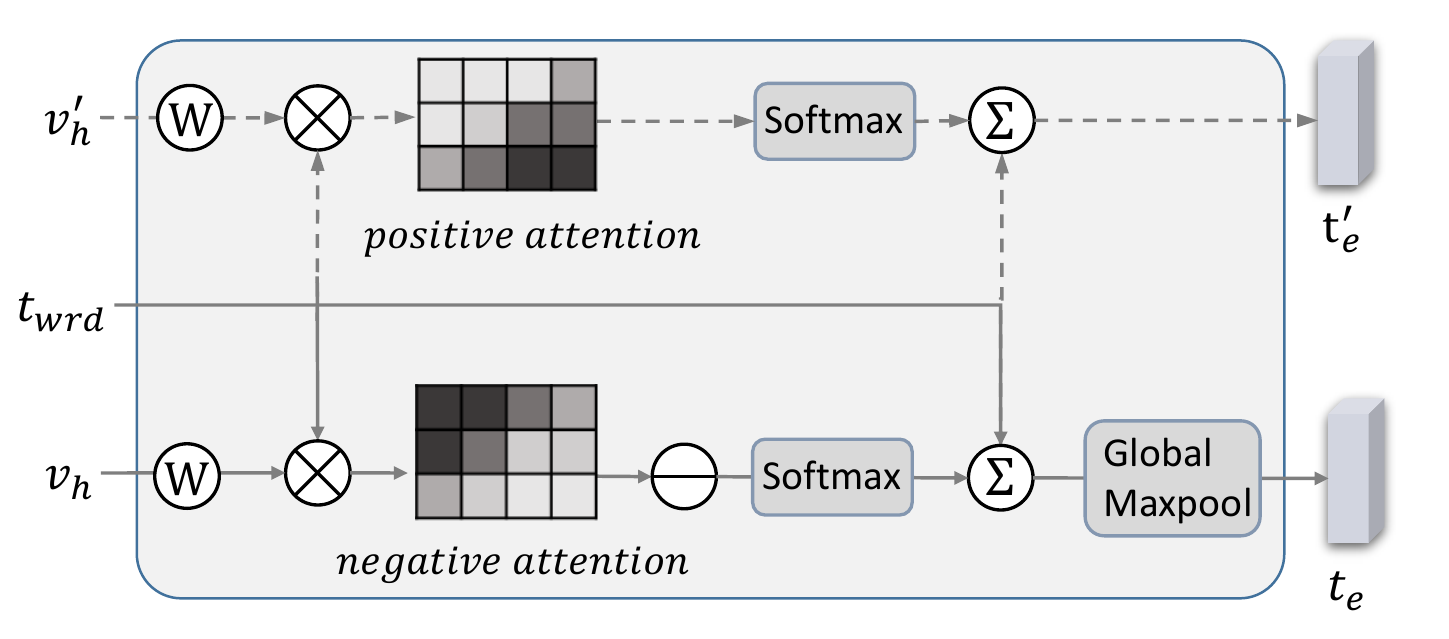}
	\caption{Detailed diagram of the Dual Multimodal Attention. The dotted path is auxiliary, while the bold one is the inpainting path.  $W$ denotes an 1x1 convolution layer.}
	\label{fig:attention}
	\vspace{-15pt}
\end{figure}
The guidance provided by the text lies in the semantics about the missing region. Therefore, the model needs to compare the input text with the corrupted image, and highlight the words that do not match the image. In this matching task, there are only negative pairs that consists of $\mathbf{I}_m$ and $\mathbf{T}$. It would be hard to learn the match function if there are no positive pairs. Therefore, we propose a dual multimodal attention mechanism that takes advantage of the dual structure and uses $\mathbf{I}_c$ and $\mathbf{T}$ to constitute positive pairs. The mechanism extracts complementary features from the word embedding in two paths by reciprocal attention. Our basic assumption for this design is that $\mathbf{I}_m$ and $\mathbf{I}_c$ are complementary. The assumption is robust in the text-guided image inpainting scenario: if $\mathbf{I}_m$ and $\mathbf{I}_c$ are not complementary, we can solely use visual features $v_h$ to infer the missing region, and do not need to specifically use the text features.

The detailed computation diagram of this mechanism is shown in Fig.~\ref{fig:attention}. The visual representations of the two paths are firstly transformed through an 1x1 convolution. Then in the auxiliary path,  the attention weights between $v_h$ and word representations $t_{wrd}$ are computed by multiplication. We keep the mask of the input image and apply it to the feature map on the same position. The computing can be formulated as Eq.~\ref{eq:attn_positive}:
\begin{equation}
\label{eq:attn_positive}
s'_{i,j} = M'_iQ(v'_{hi})^Tt_{wrdj}
\end{equation}
where $s$ denotes the attention map, $M'$ is the binary mask (the value for masked pixels is 0 and elsewhere is 1), $Q(v'_h)$=$Wv'_h$, and $W$ is the 1x1 convolution filter.

In the inpainting path, the attention weights between $v_h$ and $t_{wrd}$ is calculated to highlight the semantics about $\mathbf{I}_c$ as Eq.~\ref{eq:attn_negative}. 
\begin{equation}
\label{eq:attn_negative}
s_{i,j} = -Q(v_{hi})^Tt_{wrdj} + M_i
\end{equation}
The attention is calculated reciprocally to Eq.~\ref{eq:attn_positive} by negating the multiplication results. Specifically, in $M$, the value for masked pixels is set to $-\infty$ and elsewhere to 0.

Attention maps of both paths are fed to softmax as Eq.~\ref{eq:attn_map} to get weights for the text embeddings.
\begin{equation}
\label{eq:attn_map}
\beta_{j,i} = \frac{\exp(s_{i,j})}{\sum_{i=1}^{N}\exp(s_{i,j})} 
\end{equation}
where N is the area of the feature map, $\beta$ denotes the attention weights. According to the weights, word representations in both paths are weighted and summed up as Eq.\ref{eq:attn_sim}:
\begin{equation}
\label{eq:attn_sim}
t_{ei} = \sum_{j=1}^{L}\beta_{i,j}t_{wrdj}
\end{equation}
where $L$ is the length of the sentence, $t_{e}$ are the weighted word representations. 

There is still a problem for $t_{e}$ in the inpainting path now: after multiplying the attention weights, the features in $t_e$ are distributed following position in $v_h$, so the values at the masked region in $t_e$ are still zero, making it hard to be processed by the following convolutional layers. To handle this problem, we apply a global max-pooling on $t_e$. Since the specific location of missing semantic is unknown, we then uniformly replicate the max-pooling output. 

\subsection{Inpainting Generation}
\label{sec.gen}
The inpainting generation part takes the image and text features as prior and predicts the original image. The  features extracted from text and image are combined as multimodal hidden features $h$. Then, $h$ is fed to a fusion network as the prior condition and projected to a latent space. We assume the latent space is a Gaussian distribution and use the fusion network to predict a group of parameters for the latent space. The fusion process in the inpainting path is formulated as Eq.~\ref{eq:encode_distribution}
\begin{equation}
\label{eq:encode_distribution}
\mu, \sigma = F(h) \quad\textrm{where } h = [v_h;t_e;t_{sent}]
\end{equation}
where $\mu$ and $\sigma$ are mean and variance of the predicted Gaussian distribution, $F$ denotes the fusion network, which consists of a 5-layer residual blocks with spectral normalization~\cite{Spectral}. In the auxiliary path, distribution parameters $\mu'$ and $\sigma'$ are calculated with the same process, but the fusion network $F'$ does not share weights with $F$, because $h$ and $h'$ consists of different features.

Next, a multimodal representation $r$ is obtained by sampling latent variables from the distribution and combine the variables with $h$ through a residual connection as in Eq.~\ref{eq:encode_representation},
\begin{equation}
\label{eq:encode_representation}
r = h + Gaussian(\mu,\sigma)
\end{equation}
where $r$ is the multimodal representation. In the auxiliary path, the multimodal representation $r'$ is also obtained based on the hidden representation $h$ in the inpainting path as Eq.~\ref{eq:rec_r}, because we need to keep the multimodal representation of these two paths homogeneous.
\begin{equation}
\label{eq:rec_r}
r' = h + Gaussian(\mu',\sigma')
\end{equation}

Finally, the up-sampling networks produce a synthesized image $\mathbf{I}_g$ based on the multimodal representations on the two paths. Since image inpainting task does not require predicting the remaining areas, we feed the low-level image feature $v_l$ to the generator through a high way path with short+long term attention mentioned in the preliminary section~\ref{sec.pre} to reconstruct the pixels of $\mathbf{I}_c$. The up-sampling network consists of 5-layer residual generator network shared by the two paths.

\subsection{Optimization}
\subsubsection{Objective}
In the TDANet, the inpainting path predicts the masked region based on $\mathbf{I}_m$ and $\mathbf{T}$, while the auxiliary path reconstructs the image because the information of the full images are available. Based on the visual features $v_h$, a complete distribution for $\mathbf{I}_m$ could be learnt in the auxiliary path, and be used to guide the distribution learning in the inpainting path. 

To keep the homogeneity between the latent space in the auxiliary path and inpainting path, we also feed text features to the auxiliary path, and reconstruct the image based on multimodal prior $h'$. The conditional variational lower bound of the auxiliary path is formulated as Eq.~\ref{eq:vae_r}:
\begin{equation}
\label{eq:vae_r}
\begin{aligned}
\log p(\mathbf{I}_c|h') &\ge - \text{KL}(q_\psi(\mathbf{z}|\mathbf{I}_c,h')||p_\phi(\mathbf{z}|h')) \\
&+ \mathbb{E}_{q_\psi(\mathbf{z}|\mathbf{I}_c,h')}[\log p_\theta(\mathbf{I}_c|\mathbf{z})]
\end{aligned}
\end{equation}
where $\mathbf{z}$ is the latent vector,  $q_\psi(\cdot|\cdot)$ denotes posterior sampling function, $p_\phi(\cdot|\cdot)$ denotes the conditional prior, $p_\theta(\cdot|\cdot)$ denotes the likelihood, with $\psi$, $\phi$, and $\theta$ being the deep network parameters of their corresponding functions. 

In the inpainting path, the prior consists of the features from $\mathbf{I}_c$ and $\mathbf{T}$. The lower bound is formulated as Eq.~\ref{eq:vae_g}:
\begin{equation}
\label{eq:vae_g}
\begin{aligned}
\log p(\mathbf{I}_c|h) &\ge - \text{KL}(q_\psi(\mathbf{z}|\mathbf{I}_c,h)||p_\phi(\mathbf{z}|h)) \\
& + \mathbb{E}_{q_\psi(\mathbf{z}|\mathbf{I}_c,h)}[\log p_\theta(\mathbf{I}_c|\mathbf{z})]
\end{aligned}
\end{equation}
The inpainting quality could be optimized by maximizing the variation lower bound.
However, $q_\psi(\mathbf{z}|\mathbf{I}_c,h)$ can not be calculated directly because $\mathbf{I}_c$ is not available in the inpainting path. To solve this problem, we assume that $r$ could approximate to $r'$ after optimizing inpainting path encoders, so we have $q_\psi(\mathbf{z}|\mathbf{I}_c,h') \approx q_\psi(\mathbf{z}|\mathbf{I}_c,h)$, and Eq.~\ref{eq:vae_g} is updated as Eq.\ref{eq:vae_g_2}:
\begin{equation}
\label{eq:vae_g_2}
\begin{aligned}
\log p(\mathbf{I}_c|h) &\ge - \text{KL}(q_\psi(\mathbf{z}|\mathbf{I}_c,h')||p_\phi(\mathbf{z}|h)) \\
& + \mathbb{E}_{q_\psi(\mathbf{z}|\mathbf{I}_c,h)}[\log p_\theta(\mathbf{I}_c|\mathbf{z})]
\end{aligned}
\end{equation}
The parameters of the TDANet are learnt by maximizing these two lower bound and minimize the distance between $q_\psi(\mathbf{z}|\mathbf{I}_c,h')$ and $q_\psi(\mathbf{z}|\mathbf{I}_c,h)$.
\subsubsection{Loss Function}
We assumed in section~\ref{sec.gen} that the latent vector follows a Gaussian distribution. To optimize the variation lower bound and enforce the smoothness of the conditioning manifold space~\cite{cvaetutorial}, we maximize the Kullback-Leibler divergence term between and the standard Gaussian distribution.  For the auxiliary path, the distribution loss is formulated as Eq.\ref{eq:kl_r}:
\begin{equation}
\label{eq:kl_r}
\mathcal{L}'_{KL}= -KL(q_\psi(\mathbf{z}|\mathbf{I}_c,h')||\mathcal{N}(0,1))
\end{equation} 
As for the inpainting path, we steer the conditional prior to close to the auxiliary path posterior, which can be formulated as Eq.~\ref{eq:kl_g}:
\begin{equation}
\label{eq:kl_g}
\mathcal{L}_{KL} = -KL(q_\psi(\mathbf{z}|\mathbf{I}_c,h')||p_\phi(\mathbf{z}|h))
\end{equation} 

Next, the likelihood term $p_\theta(\mathbf{I}_c|\mathbf{z})$ need to be optimized. The likelihood term could be interpreted from both appearance and semantic aspects.

If $\mathbf{I}_g$ is close to $\mathbf{I}$ in appearance distance, the likelihood of generating $\mathbf{I}_c$ is improved. For image appearance quality, we incorporate reconstruction and adversarial losses. Three loss functions are applied as Eq.~\ref{eq:quality_loss}. 
\begin{equation}
\label{eq:quality_loss}
\mathcal{L}_{I} = ||\mathbf{I}-\mathbf{I}_g||_1 + ||D_1(\mathbf{I})-D_1(\mathbf{I}_g)||_2 + [D_2(\mathbf{I}_g)-1]^2
\end{equation}
The first term matches the per-pixel distance through $L_1$ distance. The second term is mean feature match loss~\cite{cvaetutorial}, where $D_1$ is a discriminator that returns a representation vector from the final layer. In the third term, $D_2$ indicates whether $\mathbf{I}_g$ is a real word image, the term is based on the loss in LSGAN~\cite{plu_26}, which has been proven ~\cite{2019_pluralistic} to perform better than the original GAN loss in the inpainting scenario. The networks of $D_1$ and $D_2$ are 5-layer ResNet.

To refine the semantic relation of text and image output, we adapt the DAMSM loss to our model. The match networks extract features from the generated image $\mathbf{I}_g$ through a text-image attention network and compare these features with text features. We follow the setup in ~\cite{attngan} and set $\gamma$ to 5 and formulate the loss term in Eq.~\ref{eq:match_loss}. 
\begin{equation}
\label{eq:match_loss}
\mathcal{L}_{T}  = DAMSM(t_{wrd}, \mathbf{I}_g)
\end{equation}
The loss is calculated in both paths with shared matching networks.

Finally, the total loss function could be formulated as Eq.~\ref{eq:loss}:
\begin{equation}
\label{eq:loss}
\mathcal{L} = \lambda_{KL}(\mathcal{L}'_{KL}+\mathcal{L}_{KL}) + \lambda_{I}(\mathcal{L}_{I}+\mathcal{L}'_{I}) + \lambda_{T}(\mathcal{L}_{T}+\mathcal{L}'_{T})
\end{equation}
where $\mathcal{L}_{KL} $ regularizes KL divergence of prior and posterior distribution,  $\mathcal{L}_{I}$ and $\mathcal{L}_{T}$ maximize the expectation term from image quality and text-image semantic similarity perspectives. 

\begin{table}
	\begin{center}
		\begin{tabular}{l|c|c}
			\hline
			Dataset & CUB & COCO \\ 
			\hline
			\noalign{\smallskip}
			Mask Area (Avg) &  34.59\% & 19.62\% \\
			Vocabulary size  & 5,450 & 27,297 \\
			Caption length (Avg)  & 15.23 & 10.45 \\
			Objects per image (Avg) & 1 & 7.3 \\
			Captions per image & 10 & 5\\
			\hline
		\end{tabular}
	\end{center}
	\vspace{5pt}
	\caption{Statistics for object mask and related properties on CUB and COCO dataset. The average mask area of COCO is calculated with the max box.}
	\vspace{-25pt}
	\label{tab:dataset}
\end{table}

%
%
%
%
%

\begin{table*}
	\begin{center}
		\begin{tabular}{c|c|c|c|c||c|c|c||c|c|c||c|c|c}
			\hline
			\multirow{2}{*}{Dataset} &\multirow{2}{*}{Model} & \multicolumn{3}{c||}{$\ell_1^-(\%)$} & \multicolumn{3}{c||}{PSNR$^+$} & \multicolumn{3}{c||}{TV loss$^-(\%)$} & \multicolumn{3}{c}{SSIM$^+(\%)$} \\
			\cline{3-14}
			& & center & object & $||\Delta||$ & center & object & $||\Delta||$ & center & object & $||\Delta||$ & center & object & $||\Delta||$ \\
			\hline
			\multirow{3}{*}{CUB} &CSA~\cite{2019_coherent}  & 3.99 & 6.42 & 2.43 & 20.79 & 19.13 & 1.66 & 3.64 & 4.33 & 0.69 & 82.69 & 72.65 & 10.4 \\
			& PICNet~\cite{2019_pluralistic} & 3.65 & 7.28 & 3.63 & 20.96 & 18.80 & 2.16 & 3.71 & 4.12 & 0.41 & 84.51 & 70.78 & 13.73 \\
			& TDANet  & \textbf{3.53} & \textbf{4.80} & \textbf{1.27} & \textbf{21.30} & \textbf{20.89} & \textbf{0.41} & \textbf{3.55} & \textbf{3.34} & \textbf{0.21} & \textbf{84.63} & \textbf{79.16} & \textbf{5.47} \\
			\hline
			\multirow{3}{*}{COCO} &CSA~\cite{2019_coherent} & 5.07 & 8.78 & 3.71 			   &20.07 & 19.23 & 0.84 & 4.19 & 4.86 & 0.67 & 83.21 & 75.22 & 7.99\\
			& PICNet~\cite{2019_pluralistic} & 5.53 & 9.21 & 3.68 & 19.57 & 18.73 & 1.02 & 4.51 & 4.97 & 0.46 & 81.78 & 74.44 & 7.34 \\
			& TDANet  & \textbf{4.08} & \textbf{7.48} & \textbf{3.40} & \textbf{21.31} & \textbf{20.57} & \textbf{0.74}  & \textbf{4.54} & \textbf{4.20} &\textbf{0.34}& \textbf{83.87} & \textbf{76.78} & \textbf{7.09} \\
			\hline
		\end{tabular}
	\end{center}
	\vspace{5pt}
	\caption{Quantitative comparison with the state-of-the-art on CUB and COCO dataset.   $||\Delta||$ is the performance difference between the object mask and the center mask, showing the robustness of the method at different mask positions. $^-$ lower is better, $^+$ higher is better.}
	\vspace{-15pt}
	\label{tab:CUB}
\end{table*}

\section{Experiments}
\begin{figure*}
	\begin{center}
		\setlength{\tabcolsep}{0.1em}
		\begin{tabular}{cccccc}
			\centering
			\includegraphics[width=2.2cm,height=2.2cm]{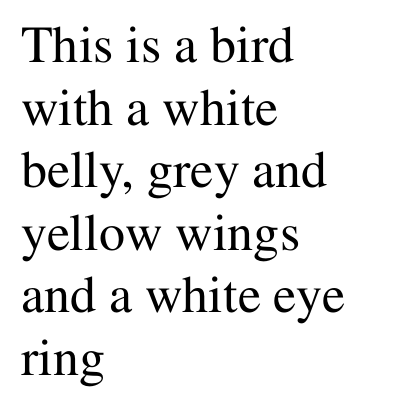}
			& 
			\includegraphics[width=2.2cm,height=2.2cm]{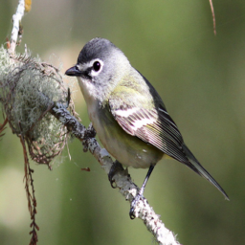} &
			\includegraphics[width=2.2cm,height=2.2cm]{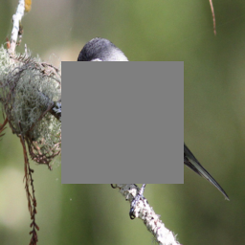} &
			\includegraphics[width=2.2cm,height=2.2cm]{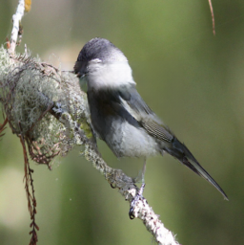} &
			\includegraphics[width=2.2cm,height=2.2cm]{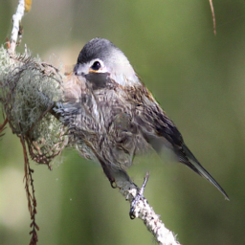} &
			\includegraphics[width=2.2cm,height=2.2cm]{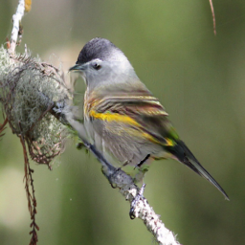} \\
			
			\includegraphics[width=2.2cm,height=2.2cm]{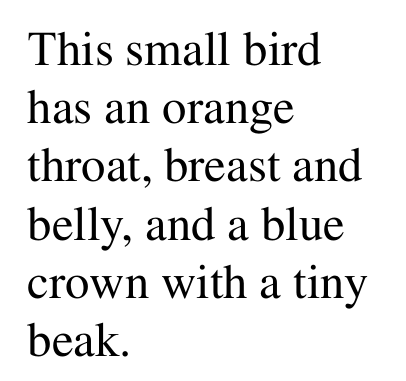}  &
			\includegraphics[width=2.2cm,height=2.2cm]{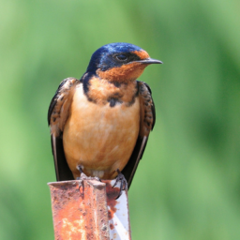} &
			\includegraphics[width=2.2cm,height=2.2cm]{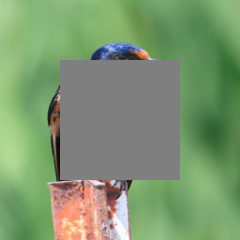} &
			\includegraphics[width=2.2cm,height=2.2cm]{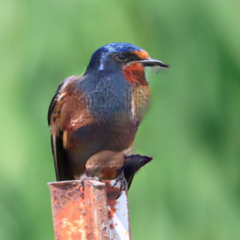} &
			\includegraphics[width=2.2cm,height=2.2cm]{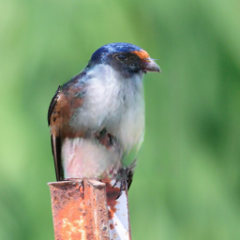} &
			\includegraphics[width=2.2cm,height=2.2cm]{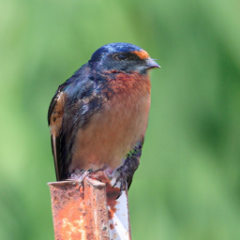} \\
			\includegraphics[width=2.2cm,height=2.1cm]{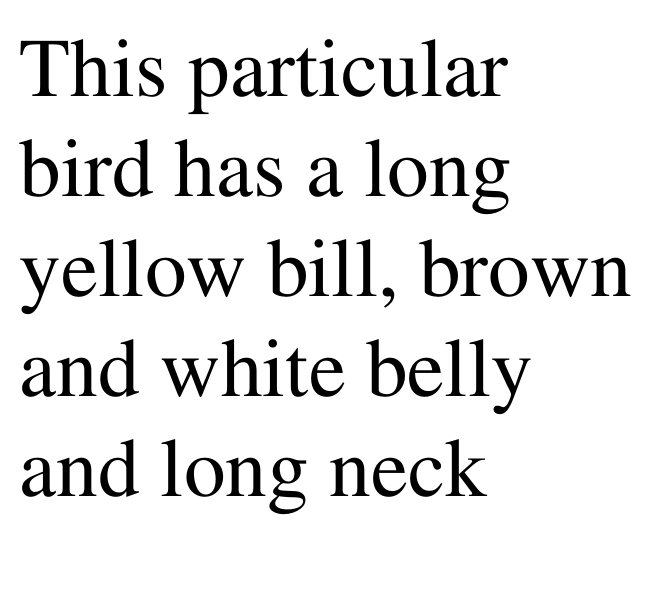} & 
			\includegraphics[width=2.2cm,height=2.2cm]{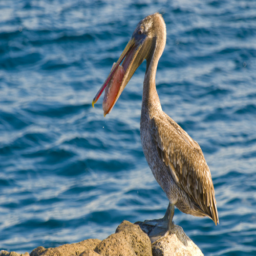} &
			\includegraphics[width=2.2cm,height=2.2cm]{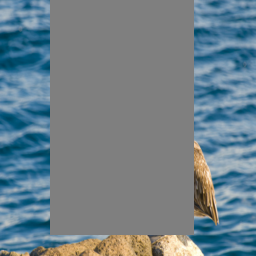} &
			\includegraphics[width=2.2cm,height=2.2cm]{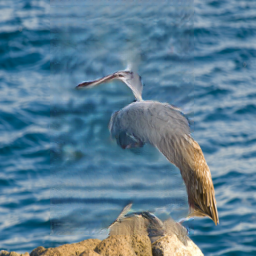} &
			\includegraphics[width=2.2cm,height=2.2cm]{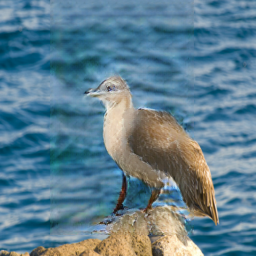} &
			\includegraphics[width=2.2cm,height=2.2cm]{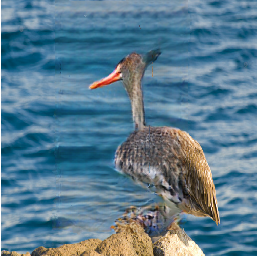}\\
			\includegraphics[width=2.2cm,height=2.2cm]{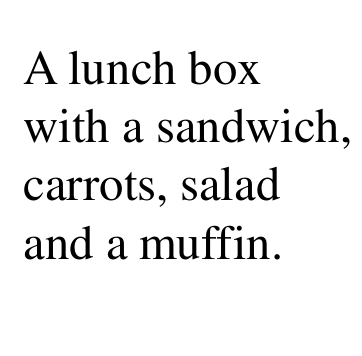}  &
			\includegraphics[width=2.2cm,height=2.2cm]{images/comparison_6_original.png} &
			\includegraphics[width=2.2cm,height=2.2cm]{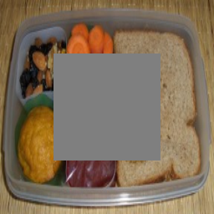} &
			\includegraphics[width=2.2cm,height=2.2cm]{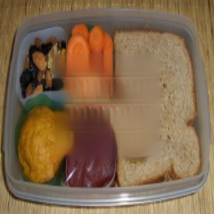} &
			\includegraphics[width=2.2cm,height=2.2cm]{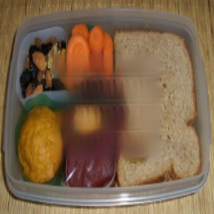} &
			\includegraphics[width=2.2cm,height=2.2cm]{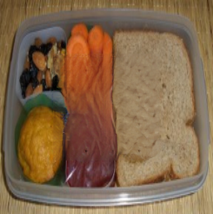}\\	
			\includegraphics[width=2.2cm,height=2.0cm]{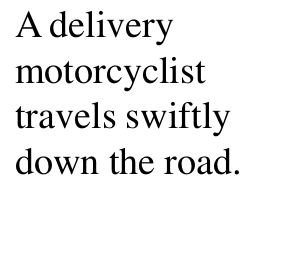}  &
			\includegraphics[width=2.2cm,height=2.2cm]{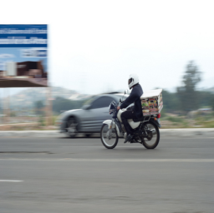} &
			\includegraphics[width=2.2cm,height=2.2cm]{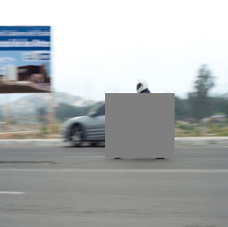} &
			\includegraphics[width=2.2cm,height=2.2cm]{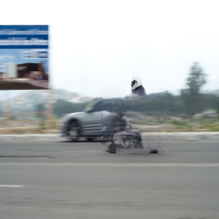} &
			\includegraphics[width=2.2cm,height=2.2cm]{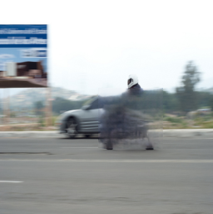} &
			\includegraphics[width=2.2cm,height=2.2cm]{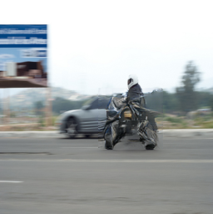} \\
			\small Text  & \small Ground truth & \small Corrupted &
			\small  CSA~\cite{2019_coherent} & \small  PICNet~\cite{2019_pluralistic} & \small  TDANet (Ours)
		\end{tabular}\\
	\end{center}
	\vspace{-10pt}
	\caption{Qualitative comparisons on CUB and COCO validation sets. (Best viewed with zoom-in)}
	\vspace{-5pt}
	\label{fig:comparison}
\end{figure*}

\subsection{Datasets}
\subsubsection{Images and Descriptions} The proposed method is evaluated on image captioning datasets CUB~\cite{cub} and COCO~\cite{coco}, with their original train, test, and validation split. The captions are used as a description for inpainting. Compared with CUB, COCO captions have a larger vocabulary but shorter and fewer sentences for each image. The properties of the two datasets are given in Table~\ref{tab:dataset}. COCO contains more instances and scenes than CUB, so learning to complete the images of coco is more challenging than CUB. 

\subsubsection{Center Mask} Following common inpainting evaluation setting~\cite{2019_gate,ma2019coarse,2019_coherent,2019_pluralistic}, a square mask taking 50\% area is constructed at the center of every image. The center mask is wildly used to compare inpainting models, but could not test their robustness at unique areas.

\subsubsection{Object Mask} Object regions are important to an image but are hard to restore. These regions are ideal test environment for unique area inpainting evaluation but have not been used yet. To evaluate our model, we construct an object mask from object detection boxes.  The COCO dataset provides dense object boxes for each image, we only select the one with the largest area, because using all masks will sometimes remove most pixels of an image and break the content. The object masks are various in size depending on the size of the objects. 

\subsection{Implementation Details}
\begin{figure*}
	\begin{center}
		\setlength{\tabcolsep}{0.15em}
		\begin{tabular}{cccccl} 
			\centering
			\includegraphics[width=2cm,height=2cm]{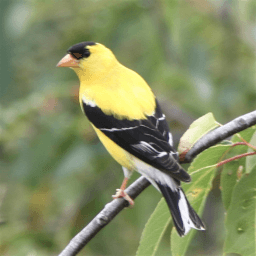} &
			\includegraphics[width=2cm,height=2cm]{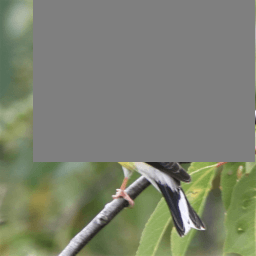} &
			\includegraphics[width=2cm,height=2cm]{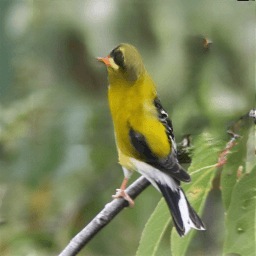} &
			\includegraphics[width=2cm,height=2cm]{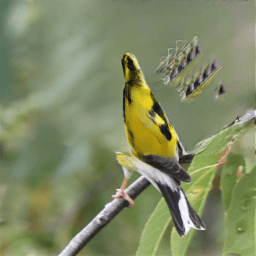} &
			\includegraphics[width=2cm,height=2cm]{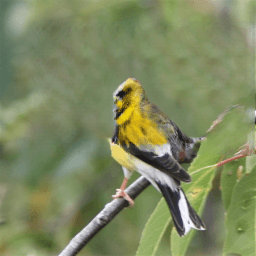} &
			\includegraphics[width=2cm,height=2cm]{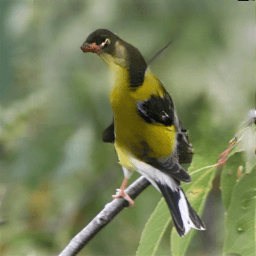} \\
			\includegraphics[width=2cm,height=2cm]{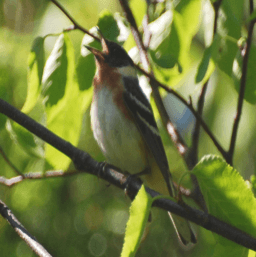} &
			\includegraphics[width=2cm,height=2cm]{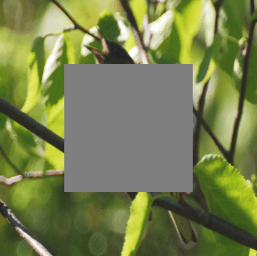} &
			\includegraphics[width=2cm,height=2cm]{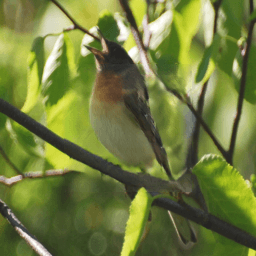} &
			\includegraphics[width=2cm,height=2cm]{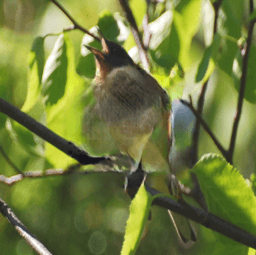}  &
			\includegraphics[width=2cm,height=2cm]{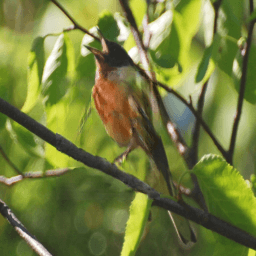} &
			\includegraphics[width=2cm,height=2cm]{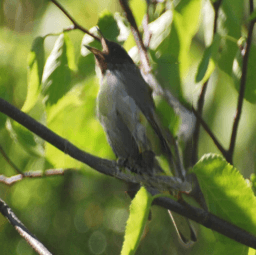}\\
			\footnotesize Original & \footnotesize corrupted &\footnotesize  TDANet & \footnotesize w/o maxpool  & \footnotesize w/o match & \footnotesize w/o dual attention \\
		\end{tabular}\\	
		
		\begin{tabular}{l}
			\small{\textit{Text 1: ``This bird has a black crown and tiny beak. the breast, throat, and nape are yellow. the wings are black with white wingbars.''}} \\
			\small{\textit{Text 2: ``The bird has brown throat, white breast, belly and abdomen, gray wings with two wingbars.''}} \\
		\end{tabular}
	\end{center}
	\vspace{-5pt}
	\caption{Qualitative cases of the ablation study.}
	\label{fig:ablation}
\end{figure*}
For images in both CUB and COCO datasets, the training images are resized to make their minimal length/width as 256 and crop the sub-image of size $256$x$256$ at the center. 
Networks are initialized with orthogonal initialization~\cite{plu_33} and trained end-to-end with a learning rate of $10^{-4}$ on Adam optimizer~\cite{plu_18}. The image-text matching networks are pre-trained as in \cite{attngan} on CUB and COCO, respectively. During training, the weights of loss function terms are set as $\lambda_{KL}=\lambda_{I}=20$, $\lambda_{T}=0.1 $. The maximum length of text sequence is set to 128 with zero padding. Experiments are conducted on Ubuntu 18.04 system, with i7-9700K 3.70GHz CPU and 11G NVIDIA RTX2080Ti GPU. 
We compare the proposed TDANet with two state-of-the-art image inpainting methods PICNet~\cite{2019_pluralistic} and CSA~\cite{2019_coherent} based on their official source codes
\footnote{
	\url{https://github.com/KumapowerLIU/CSA-inpainting}\\
	\url{https://github.com/lyndonzheng/Pluralistic-Inpainting}
}. 

\subsection{Quantitative Results}
Inpainting quality depends on the authenticity of the filled area and its continuity with the surrounding images.  Existing quantitative metrics for inpainting models could only evaluate  these characteristics roughly \cite{2018_plu_42_contextualattention,2019_gate}. To compare with other models, we select commonly used mean $\ell_1$ loss, peak signal-to-noise ratio (PSNR), total variation (TV), and Structural Similarity (SSIM) for quantitative comparison. We will further discuss the semantic consistency in other experiments. 

From the results in Table~\ref{tab:CUB}, the proposed TDANet outperforms compared models in all measures. The performance improvement on the object mask is more obvious than the center mask. Comparing the improvements on the two datasets, TDANet improves more on CUB than COCO, which qualifies our expectation that combining image and caption on COCO is more challenging. 
It is worth noting that the performance of all evaluated models decreases on object masked samples, while our model has the least performance degradation, proving the robustness in such unique areas.

\definecolor{atomictangerine}{rgb}{1.0, 0.6, 0.4}
\subsection{Qualitative Results}
Both inpainting quality and semantic consistency are evaluated in qualitative comparison.
Fig.~\ref{fig:comparison} shows the qualitative results on the CUB and COCO validation set. As shown in the figure, the results of all the three models on the CUB dataset are consistent with the surrounding areas but compared with the ground truth, the CSA and PICNet outputs fail to recover some special characteristics, such as the belly color and neck shape. Through careful observation, it can be found that the content filled by CSA and PICNet has similar characteristics to the neighbor pixels such as color and texture. In other words, these models are filling unique areas by borrow features from their surroundings. In comparison, the images generated by the TDANet produce these unique features as mentioned in the text, so the results look semantically consistent with the original image. On the challenging COCO dataset, all these three models are imperfect, but our model obtains a better appearance quality compared with other models. 

\begin{table}
	\begin{center}
		\begin{tabular}{lcc}
			\hline\noalign{\smallskip}
			Model & Naturalness & Semantic Consistency\\
			\noalign{\smallskip}
			\hline
			\noalign{\smallskip}
			Ground Truth & 1.208 & 1.363 \\
			\hline
			\noalign{\smallskip}
			CSA  & 2.981 & 3.060 \\
			PICNet  & 3.178 & 3.469 \\
			TDANet  & \textbf{2.531} & \textbf{2.106}\\
			\hline
		\end{tabular}
	\end{center}
	\vspace{5pt}
	\caption{Numerical ranking score of the user study. The lower the score, the better the performance.}
	\vspace{-25pt}
	\label{tab:userstudy}
\end{table}
\begin{figure}
	\centering
	\includegraphics[width=0.7\linewidth, height=0.5\linewidth]{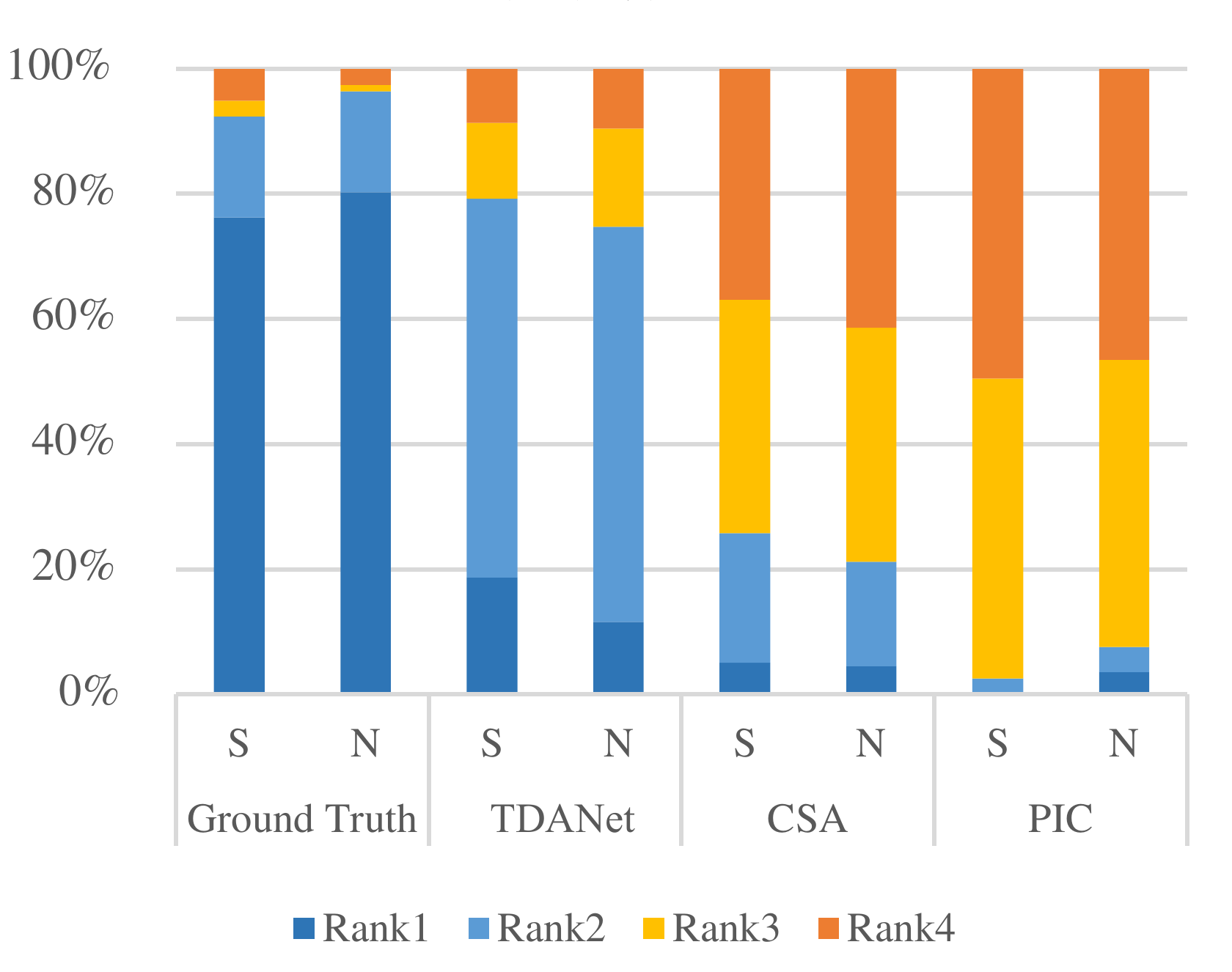}
	\vspace{-5pt}
	\caption{Ranking score distribution of the user study. "S" means semantic consistency score, "N" means naturalness.}
	\vspace{-10pt}
	\label{fig:userstudy}
\end{figure}
\subsection{User Study}
A ranking game was designed to further quantify the qualitative comparison from the human perspective. We randomly collected 100 images in center masks from the CUB validation dataset (see details in supplementary files). For each image, a tuple consisting of (1) Ground truth, (2) Our model, (3) PICNet, (4) CSA model is prepared. We randomly shuffled the four samples in each tuple and recruited 20 volunteers to rank them according to naturalness and the semantic consistency with the text description. After the test, we computed the average of the ranking score for the four groups. 

The results are shown in Table~\ref{tab:userstudy}. According to the results, our model performs better than other models in terms of realistic, and significantly higher in semantic consistency. We also counted the rank score distribution of each model in Fig.~\ref{fig:userstudy}. Compared with other models, TDANet results ranks first and second more often than the other two models, and third and fourth less often. We also found that sometimes TDANet could mislead the tester and get a better rank than the original image. We also noticed that TDANet gets ranks better in semantic consistency than naturalness, which means sometimes the model grasps the key information from the text but can not generate the content perfectly due to the limitation of the generator.

\subsection{Ablation Study}
Four models without each component were trained using the same super parameters and epochs. As shown in Table~\ref{tab:ablation}, w/o the auxiliary path and dual multimodal attention lead to the most obvious performance decrease on every metric. w/o the global max-pooling layer and replication of text features mainly lead to appearance metrics decrease, which is reflected by TV loss and SSIM. We have noticed that removing the match loss leads to appearance metrics decrease while pixel metrics increases. 

To further inspect the effect of the components, we compared their result qualitatively; the cases are given in Fig~\ref{fig:ablation}. The quantitative ablation comparison shows that although w/o the match loss leads to higher pixel metrics, its output looks worse and fails to allocate characters precisely.

\begin{table}
	\begin{center}
		\begin{tabular}{lcccc}
			\hline\noalign{\smallskip}
			Model & $\ell_1^-$ & PSNR$^+$ & TV loss$^-$ & SSIM$^+$\\
			\noalign{\smallskip}
			\hline
			\noalign{\smallskip}
			TDANet  & 4.80 & 20.89 & 3.34 & 79.16\\
			\hline
			\noalign{\smallskip}
			w/o match loss  & 4.81 & 20.93 & 3.48 & 78.58\\
			w/o maxpool  & 4.71 & 20.83 & 3.48 & 78.85\\
			w/o multimodal attention   & 5.77 & 20.14 & 3.78 & 74.31 \\
			\hline
		\end{tabular}
	\end{center}
	\vspace{5pt}
	\caption{Numerical results of the ablation study.}
	\vspace{-20pt}
	\label{tab:ablation}
\end{table}

\subsection{Controllability Evaluation and Manipulation Extension}
The text guidance not only benefits inpainting quality but also allows changing the model output by providing various descriptions. The interactive manner enables more application scenarios, such as manipulation. Here we show the controllability and extension to image manipulation at the same time. The manipulation has three steps: First, select the region to be changed and mask it. Second, write a sentence describing the desired image after manipulation. Third, feed the masked image and sentence to the TDANet and get the result. For each image, we present two results when two different sentences are entered. 

\begin{figure}
	\begin{center}
		\setlength{\tabcolsep}{0.1em}
		\begin{tabular}{cccc}
			\centering
			\includegraphics[width=2cm,height=2cm]{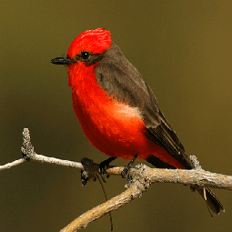} & 
			\includegraphics[width=2cm,height=2cm]{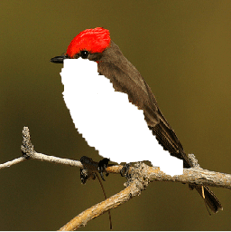} &
			\includegraphics[width=2cm,height=2cm]{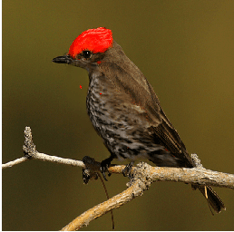} &
			\includegraphics[width=2cm,height=2cm]{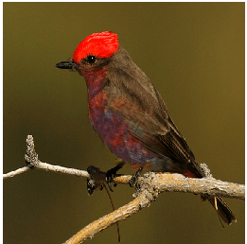}  \\
			\footnotesize{Input} & \footnotesize{Mask} & \footnotesize{Result (a)} & \footnotesize{Result (b)} \\
			\multicolumn{4}{l}{\footnotesize{\textit{Text (a): ``The bird has \textbf{black and white spotted breast} a red hat and grey wings.''}}} \\
			\multicolumn{4}{l}{\footnotesize{\textit{Text (b): ``The bird has a {\color{purple}purple breast} a red hat and grey wings.''}} } \\
			\includegraphics[width=2cm,height=2cm]{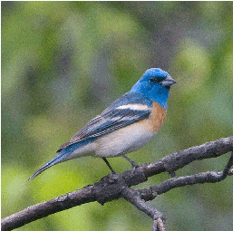} & 
			\includegraphics[width=2cm,height=2cm]{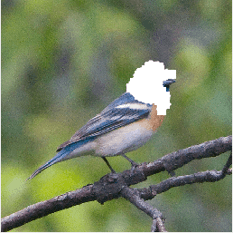} &
			\includegraphics[width=2cm,height=2cm]{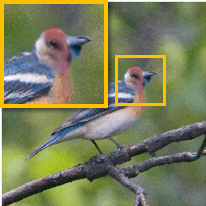} &
			\includegraphics[width=2cm,height=2cm]{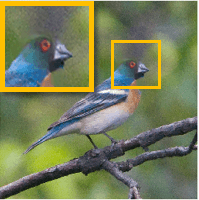} \\
			\footnotesize{Input} & \footnotesize{Mask} & \footnotesize{Result (a)} & \footnotesize{Result (b)} \\
			\multicolumn{4}{l}{\footnotesize{\textit{Text (a): ``A bird with a white breast and a {\color{atomictangerine} brown crown}.''}}} \\
			\multicolumn{4}{l}{\footnotesize{\textit{Text (b): ``a bird with a white breast blue head and {\color{red}red eyes}. }''}} \\
			\includegraphics[width=2cm]{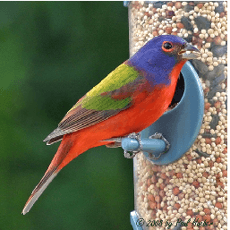} & 
			\includegraphics[width=2cm]{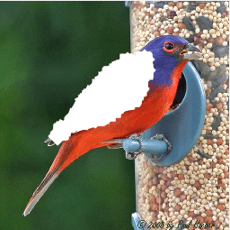} &
			\includegraphics[width=2cm]{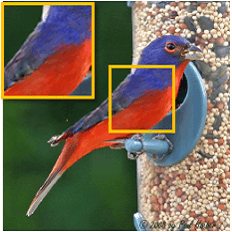} &
			\includegraphics[width=2cm]{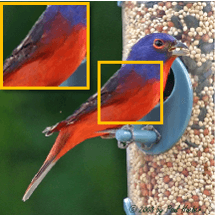} \\
			\footnotesize{Input} & \footnotesize{Mask} & \footnotesize{Result (a)} & \footnotesize{Result (b)} \\
			\multicolumn{4}{l}{\footnotesize{\textit{Text (a): ``A distinct red bird with a blue head {\color{blue} blue wings} and a red eyering.''}}} \\
			\multicolumn{4}{l}{\footnotesize{\textit{Text (b): ``A distinct red bird with a blue head and {\color{red} red wings}.''}}} \\						
		\end{tabular}
	\end{center}
	\vspace{-5pt}
	\caption{TDANet with different guidance texts. }
	\vspace{-15pt}
	\label{fig:edit}
\end{figure}

Figure~\ref{fig:edit} shows the image manipulation results. The first group is color variation by masking the belly and giving sentences to describe the bird with different colors. The second group shows the edit of different parts by providing sentences describing different organs. In the third case, we selected an area between blue and red colors and command the model to fill in one of the neighboring colors. The result shows the output was not affected by the color of adjacent areas. The experiment presents a new possible formulation of image manipulation, and shows that pluralistic results could be generated by controlling the guidance text.

\section{Conclusion}
We have presented a text-guided inpainting scheme that combines the features of the image and descriptive text as the generation condition. We proposed a dual multimodal attention mechanism to exploit semantic features about the masked region from the descriptive text. What's more, we introduced an image-text matching based loss to improve the semantic consistency between inpainting output and the text. The experimental results demonstrated that the proposed TDANet outperformed compared models in subjective and objective comparisons, and the model outputs are semantically consistent with the guidance text. 

Several future directions and improvements could be considered. Our main objective was to show the potential of text-guided image inpainting. We found that with simple encoders, the semantics were well extracted; future work might consider more complex architectures to refine the representations. Performance on the COCO dataset is still limited. External visual knowledge about concepts and data augmentation will improve inpainting quality.

\section*{Acknowledgments}
We would like to thank Joanna Siebert for her helpful feedback. This work is supported by Natural Science Foundation of China (Grant No. 61872113), Strategic Emerging Industry Development Special Funds of Shenzhen (Grant No.XMHT20190108009), the Tencent Group and Science and Technology Planning Project of Shenzhen (Grant No.JCYJ20190806112210067).

\bibliographystyle{ACM-Reference-Format}
\bibliography{sample-base}

\end{document}